\documentclass[sigconf]{acmart}

\AtBeginDocument{%
  }

\setcopyright{acmlicensed}
\copyrightyear{2018}
\acmYear{2018}
\acmDOI{XXXXXXX.XXXXXXX}
\acmConference[Conference acronym 'XX]{Make sure to enter the correct
  conference title from your rights confirmation email}{June 03--05,
  2018}{Woodstock, NY}
\acmISBN{978-1-4503-XXXX-X/2018/06}

\usepackage{amsmath,amsfonts,bm}

\def\eqref#1{equation~\ref{#1}}
\def\1{\bm{1}}

\DeclareMathAlphabet{\mathsfit}{\encodingdefault}{\sfdefault}{m}{sl}
\SetMathAlphabet{\mathsfit}{bold}{\encodingdefault}{\sfdefault}{bx}{n}

\usepackage{xspace}
\usepackage{booktabs}
\usepackage{graphicx}
\usepackage{multicol, multirow, xcolor, threeparttable}
\usepackage{gensymb}
\usepackage{hyperref}
\usepackage{url}
\usepackage{makecell}
\usepackage{array}
\usepackage{tabularx}
\usepackage{listings}
\usepackage{algorithm}
\usepackage{algpseudocode}

\def\0{\mathbf{0}}

\def \X {\mathbf{X}}
\def \Y {\mathbf{Y}}
\def \Z {\mathbf{Z}}

\begin{document}


\title{Integrating Weather Foundation Model and Satellite to Enable Fine-Grained Solar Irradiance Forecasting
% Multimodal Integration of Weather Foundation Model and Satellite Imagery for Operational Solar Irradiance Forecasting
% Baguan-solar: Beyond Naïve Multimodal Fusion with Hierarchical Weather–Satellite Integration for Fine-Grained Irradiance Forecasting
%Baguan-solar: Fine-Grained Operational Solar Irradiance Forecasting via Multimodal Fusion
}

\author{Ziqing Ma}
\authornote{Authors contributed equally to this work.}
\email{maziqing.mzq@alibaba-inc.com}
\affiliation{%
  \institution{DAMO Academy, Alibaba Group}
  %\country{China}
  % \streetaddress{P.O. Box 1212}
  \city{Hangzhou}
  \state{Zhejiang}
  \country{China}
  % \postcode{}
}

\author{Kai Ying}
\authornotemark[1]
\email{yingkai.ying@alibaba-inc.com}
\affiliation{%
  \institution{DAMO Academy, Alibaba Group}
  %\country{China}
  % \streetaddress{P.O. Box 1212}
  \city{Hangzhou}
  \state{Zhejiang}
  \country{China}
  % \postcode{}
}
\affiliation{%
  \institution{College of Computer Science, Zhejiang University of Technology}
  \city{Hangzhou}
  \state{Zhejiang}
  \country{China}
}

\author{Xinyue Gu}
\authornotemark[1]
% \authornote{Corresponding authors.}
% \authornotemark[2]
\email{guxinyue.gxy@alibaba-inc.com}
\affiliation{%
  \institution{DAMO Academy, Alibaba Group}
  %\country{China}
  % \streetaddress{P.O. Box 1212}
  \city{Hangzhou}
  \state{Zhejiang}
  \country{China}
  \postcode{}
}

\author{Tian Zhou}
\authornotemark[1]
\email{tian.zt@alibaba-inc.com}
\affiliation{%
  \institution{DAMO Academy, Alibaba Group}
  \city{Hangzhou}
  \state{Zhejiang}
  \country{China}
  \postcode{}
}

\author{Tianyu Zhu}
\authornotemark[1]
\email{yunrui.zty@alibaba-inc.com}
\affiliation{%
  \institution{DAMO Academy, Alibaba Group}
  \city{Hangzhou}
  \state{Zhejiang}
  \country{China}
  \postcode{}
}

\author{Haifan Zhang}
% \authornote{Authors contributed equally to this research.}
\email{zhanghaifan.zhf@alibaba-inc.com}
\affiliation{
  \institution{DAMO Academy, Alibaba Group}
  \city{Hangzhou}
  \state{Zhejiang}
  \country{China}
}

\author{Peisong Niu}
\email{niupeisong.nps@alibaba-inc.com}
\affiliation{
  \institution{DAMO Academy, Alibaba Group}
  \city{Hangzhou}
  \state{Zhejiang}
  \country{China}
}

\author{Zheng Wang}
\orcid{https://orcid.org/0000-0002-6753-6569}
\email{zhengwang@zjut.edu.cn}
\affiliation{%
  \institution{College of Computer Science, Zhejiang University of Technology}
  \city{Hangzhou}
  \state{Zhejiang}
  \country{China}
}

 \author{Cong Bai}
\orcid{https://orcid.org/0000-0002-6177-3862}
\email{congbai@zjut.edu.cn}
\affiliation{%
  \institution{College of Computer Science, Zhejiang University of Technology}
  \city{Hangzhou}
  \state{Zhejiang}
  \country{China}
}

\author{Liang Sun}
\email{liang.sun@alibaba-inc.com}
% \authornote{Corresponding authors.}
\affiliation{
  \institution{DAMO Academy, Alibaba Group}
  \city{Bellevue}
  \country{USA}
}

% \author{Peisong Niu}
% \authornote{Authors contributed equally to this research.}
% \email{niupeisong.nps@alibaba-inc.com}
% % \orcid{1234-5678-9012}
% \affiliation{%
%   \institution{DAMO Academy, Alibaba Group}
%   %\country{China}
%   % \streetaddress{P.O. Box 1212}
%   \city{Hangzhou}
%   % \state{Ohio}
%   \country{China}
%   % \postcode{}
% }

\renewcommand{\shortauthors}{xxx et al.}
%% No italics, no superscripts
%% Use footnote or author note to identify equal contribution and/or contact author info

% \affiliation{%

% \affiliation{%



\begin{abstract}
% 先简单写一个后面再重构，正文一共8页
Accurate day-ahead solar irradiance forecasting is essential for integrating solar energy into the power grid. However, it remains challenging due to the pronounced diurnal cycle and inherently complex cloud dynamics. Current methods either lack fine-scale resolution (e.g., numerical weather prediction, weather foundation models) or degrade at longer lead times (e.g., satellite extrapolation). We propose Baguan-solar, a two-stage multimodal framework that fuses forecasts from Baguan, a global weather foundation model, with high-resolution geostationary satellite imagery to produce 24-hour irradiance forecasts at kilometer scale. Its decoupled two-stage design first forecasts day-night continuous intermediates (e.g., cloud cover) and then infers irradiance, while its modality fusion jointly preserves fine-scale cloud structures from satellite and large-scale constraints from Baguan forecasts. Evaluated over East Asia using CLDAS as ground truth, Baguan-solar outperforms strong baselines (including ECMWF IFS, vanilla Baguan, and SolarSeer), reducing RMSE by 16.08\% and better resolving cloud-induced transients. An operational deployment of Baguan-solar has supported solar power forecasting in an eastern province in China, since July 2025. Our code is accessible at https://github.com/DAMO-DI-ML/Baguan-solar.git.

\end{abstract}


 % sub-seasonal to seasonal forecasting, regional forecasting.



%     \centering
%     \includegraphics[width=0.8\textwidth]{figures/intro.pdf}
%     \vskip -0.1in
%     \caption{A brief review. 
%     (a) Pre-training can effectively alleviate overfitting.
%     % This demonstrates that our pre-training method effectively alleviate the problem of overfitting. 
%     (b) A proper task difficulty leads to better performance.
%     % Simpler pre-training tasks tend to facilitate smoother transitions to downstream tasks and result in superior performance. 
%     (c) Attention maps of Siamese MAE and MAE. (d) Downstream tasks for \NAME through fine-tuning.} 
%     \label{fig:intro}

\maketitle

% \begin{abstract}
% The abstract paragraph should be indented 1/2~inch (3~picas) on both left and right-hand margins. Use 10~point type, with a vertical spacing of 11~points. The word \textsc{Abstract} must be centered, in small caps, and in point size 12. Two line spaces precede the abstract. The abstract must be limited to one paragraph.
% \end{abstract}

\section{Introduction}

\begin{figure*}[t!]
    \centering
    \includegraphics[width=1.0\textwidth]{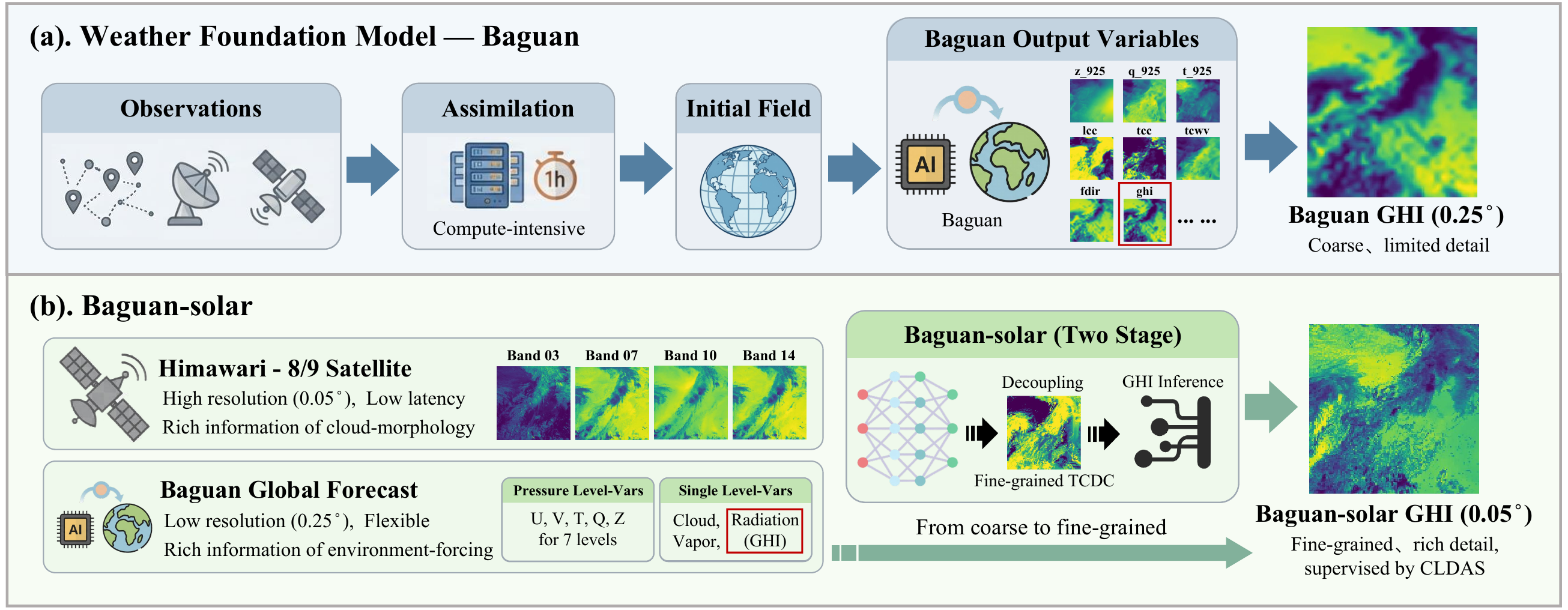}
    \caption{From coarse Baguan forecasts to fine-grained GHI: the Baguan-solar overall framework.}
    \label{fig:framework}
    \vspace{-1.0em}
\end{figure*}

% The global transition towards sustainable energy systems has positioned solar power as a cornerstone of the future electricity grid. However, the inherent variability of solar irradiance, primarily driven by cloud dynamics, poses significant challenges to grid stability and operational efficiency~\cite{aReviewOfSolarForecasting}. 
The inherent variability of solar irradiance, driven largely by cloud dynamics, presents a major challenge to integrating solar power into the electricity grid, impacting both its stability and operational efficiency~\cite{aReviewOfSolarForecasting}.
Accurate day-ahead (24~h) forecasting of surface solar irradiance, typically quantified as Global Horizontal Irradiance (GHI), is therefore essential to enable the large-scale, reliable, and cost-effective integration of solar energy into modern power systems~\cite{ReviewSystemPlanningHighSolarPower}. 

Despite its importance, day-ahead GHI forecasting remains difficult for two fundamental reasons. First, GHI exhibits a pronounced diurnal cycle: it is strictly zero at night and increases rapidly after sunrise, introducing strong %nonstationarity and 
discontinuities around day--night transitions~\cite{aReviewOfSolarForecasting}. Second, accurate forecasts must simultaneously capture fine-scale cloud morphology (which drives sharp local irradiance fluctuations) and remain skillful at longer lead times, where cloud motion errors accumulate and cloud evolution involves formation and dissipation rather than pure advection~\cite{FusionSF}.

Currently, mainstream operational and research solutions for GHI forecasting can be broadly grouped into three categories: (i) physics-based numerical weather prediction (NWP), (ii) data-driven weather foundation models (WFMs), and (iii) satellite-based extrapolation methods. NWP models solve the governing physical equations of atmospheric dynamics and radiation, offering strong physical consistency and reliable long-horizon predictability. However, they are computationally expensive and often struggle to resolve the fine-scale, rapidly evolving cloud processes that dominate local irradiance variability~\cite{solarseer, lima2016forecast}. In addition, they also suffer from the so called ``spin-up'' problem, making it often inaccurate in the early-hour prediction~\cite{warner2023assessing}.

%NWP models, which solve physical equations governing atmospheric dynamics, have traditionally been the backbone of meteorological forecasting. While they offer high physical fidelity, their computational cost is prohibitive, and they often struggle to resolve the fine-scale, rapidly evolving cloud processes that dominate irradiance fluctuations at the surface.

Recent advances in WFMs present a promising alternative to conventional NWP systems. WFMs such as Pangu‑Weather~\cite{pangu_nature}, FuXi~\cite{Fuxi_nature}, GraphCast~\cite{graphcast},  and Baguan~\cite{Baguan} have demonstrated forecasting skill surpassing NWP models at medium ranges, while drastically reducing computational expense. These models learn to predict the evolution of global atmospheric fields %directly from reanalysis data, 
typically at $0.25^\circ$.
However, current WFMs remain suboptimal for solar energy applications due to two key limitations. First, they are primarily trained to optimize conventional meteorological variables (e.g., wind, temperature, humidity), yet many lack outputs for irradiance or cloud-specific parameters that are essential for solar forecasting, as summarized in Appendix Table~\ref{app:tab:compare_wfms}. Second, their native spatial resolution is too coarse to adequately resolve mesoscale cloud structures. %, which critically influence local GHI variability. 
Furthermore, increasing the spatial resolution leads to a dramatic growth in computational cost, making high-resolution global WFMs impractical in real-world deployment~\cite{mukkavilli2023ai}.

In parallel, satellite extrapolation based methods have made compelling progress in high-resolution solar nowcasting~\cite{solarseer,FusionSF,CrossViVit}. Geostationary platforms, such as Himawari‑8/9~\cite{Intro-to-himawari8/9}, provide multi-spectral imagery at kilometer-scale resolution and minute-scale cadence, offering a rich description of cloud morphology, cloud-top temperature, and moisture structure. Models like SolarSeer~\cite{solarseer} leverage such satellite imagery to forecast GHI over large domains (e.g., CONUS) at 5 km resolution, achieving substantial speed-ups while narrowing the error gap. At longer lead times (12--24~h), satellite-only methods become a mere extrapolation problem, ignoring atmospheric dynamics in different layers and leading to degraded performance. %Yet, purely satellite-driven approaches face fundamental limitations at longer lead times (e.g., 12–24 h): cloud evolution degenerates into an extrapolation problem, and the lack of explicit large-scale dynamical information leads to degraded performance as the forecast horizon increases.

These observations highlight a critical gap: existing approaches either (i) provide physically rich but coarse and computationally expensive forecasts (NWP, WFMs), or (ii) offer high spatial resolution and low latency but rely solely on satellite extrapolation, which limits longer-horizon skill. However, a seamless integration of WFMs with satellite observations for solar irradiance forecasting remains a recognized challenge in the field. % {\color{red}Moreover, No existing system effectively integrates WFMs with satellite observations (for solar irradiance forecast?).} %there is currently no widely available forecasting system explicitly designed to combine WFMs with high-resolution satellite observations. 
To bridge this gap, we propose Baguan‑solar, a two-stage, multimodal framework that integrates WFMs forecasts with satellite imagery for fine-grained solar irradiance forecasting. The overall framework is illustrated in Figure~\ref{fig:framework}. To handle the pronounced diurnal cycle, in which GHI changes abruptly around sunrise and sunset, Baguan-solar employs a decoupled two-stage design. The first stage targets intermediate variables like total cloud cover and future satellite imagery, ensuring consistent modeling across day and night. 
%The first stage explicitly predicts intermediate, continuous variables such as total cloud cover and future satellite imagery, which are well-defined both day and night. 
These outputs then serve as inputs to the second stage, which infers GHI by combining them with clear-sky GHI and meteorological context. To jointly preserve fine-scale cloud morphology while maintaining day-ahead skill, Baguan-solar further adopts a modality-fusion design that leverages high-resolution satellite observations to capture local cloud structures and uses Baguan forecasts~\cite{Baguan} to provide large-scale dynamical and thermodynamical constraints that guide longer-horizon cloud evolution.

Baguan-solar is comprehensively evaluated over East Asia using the China Meteorological Administration’s Land Data Assimilation System (CLDAS)~\cite{Shi2013CLDAS} as ground truth. CLDAS provides a spatially and temporally consistent analysis by assimilating dense surface observations and satellite retrievals, making it a superior reference for GHI over East Asia compared with reanalysis datasets such as ERA5. %, due to its finer spatial resolution and enhanced regional accuracy enabled by a denser in-situ observation network. 
Evaluated on CLDAS, Baguan-solar outperforms a range of baselines, including vanilla Baguan~\cite{Baguan}, ECMWF Integrated Forecasting System (IFS), SolarSeer~\cite{solarseer}. Under our operational setting, Baguan-solar reduces GHI RMSE by approximately 16.08\% relative to the strongest baseline, while substantially improving the representation of rapid irradiance transients associated with passing clouds. % Moreover, an ``idealized'' variant of Baguan-solar driven by ERA5 instead of Baguan forecasts yields only marginal gains over the operational version, suggesting that modern AI weather models already provide sufficiently accurate dynamical context for high-resolution solar forecasting.
In summary, the contributions of this work are fourfold:
\vspace{-0.5em}
\begin{enumerate}

    \item We propose Baguan-solar, a multimodal fusion framework that integrates Baguan forecasts with geostationary satellite imagery to produce fine-grained day-ahead (24~h) GHI forecasts.
    \item We design a decoupled two-stage Swin Transformer that first forecasts day--night continuous cloud-related intermediates and then infers GHI. % using these outputs together with clear-sky irradiance and meteorological context.
    \item We conduct comprehensive experiments over East Asia using CLDAS as ground truth, demonstrating that Baguan-solar consistently outperforms strong baselines (including Baguan, ECMWF IFS, and satellite-based methods). %) and improves the representation of rapid irradiance transients caused by passing clouds.
    \item We demonstrate operational deployment of Baguan-solar in an online forecasting system, where it runs hourly using the latest Baguan forecasts and satellite data as input to support real-world solar power forecasting.

    % \item A hierarchical multimodal fusion framework that integrates global AI weather forecasts and geostationary satellite imagery for fine-grained, day-ahead solar irradiance forecasting.
    % \item A two-stage Swin Transformer architecture that explicitly models cloud evolution and then infers GHI, moving beyond naïve multimodal fusion and simple cloud extrapolation.
    % \item Comprehensive empirical evaluation over East Asia, demonstrating that Baguan‑Solar delivers faster, more accurate, and more scalable 24‑hour GHI forecasts than state-of-the-art baselines, thereby providing a practical tool for grid-scale solar power integration.
    % \item We demonstrate operational deployment of Baguan-solar in an online forecasting system, where it runs hourly using the latest Baguan forecasts as input to support real-world solar power forecasting.
\end{enumerate}

\section{Related Work}

\subsection{WFMs for Irradiance Forecasting}

WFMs have progressed rapidly in recent years. Existing WFMs can be broadly categorized into two architectural paradigms: graph-based and transformer-based models. Graph-based approaches~\cite{graphcast, lang2024aifs} naturally accommodate Earth's geometry and enabling flexible spatial discretization. In contrast, transformer-based approaches~\cite{Nguyen2023ClimaXAF,pangu_nature,Fuxi_nature,chen2023fengwu} typically tokenize gridded meteorological fields and leverage attention mechanisms for spatiotemporal modeling. Beyond architectural choices, Baguan~\cite{Baguan} adopts a pre-training--fine-tuning pipeline to mitigate overfitting under limited real-world data. However, most existing WFMs~\cite{graphcast,lang2024aifs,Fuxi_nature,pangu_nature,chen2023fengwu,Nguyen2023ClimaXAF,pathak2022fourcastnet} do not natively support solar irradiance forecasting. To the best of our knowledge, only Baguan and FuXi-2.0~\cite{fuxi2.0} provide irradiance-related outputs. Moreover, although WFMs like Baguan produce surface irradiance fields, their predictions are not specifically optimized for solar-energy applications and remain restricted to a coarse $0.25^\circ$ spatial resolution. % In contrast, Baguan-solar is the first method to adapt a global weather foundation model for day-ahead solar irradiance forecasting at 0.05\textdegree resolution, enabling substantially finer-grained irradiance fields.
% 目前写的这一版介绍了全球大模型A、B、C。。。分别做了什么，缺少和文章故事的结合。需要加一些串联的段。可以考虑这样写：全球大模型A、B、C大部分都没有输出辐照，有的能输出辐照，但是没有细致优化过。而且全球大模型只能输出0.25度的，很粗糙，我们是第一个做到0.05度的。这个逻辑在solarseer Table1 里也有体现，可以参考。

\subsection{From WFMs to Downstream Irradiance Products}
Recent studies demonstrate the potential of building downstream irradiance forecasting applications on top of WFMs. 
Huang et al. propose FuXi-RTM~\cite{FuxiRTM}, which couples FuXi~\cite{Fuxi_nature} with a fixed radiative transfer model to enforce radiative-transfer consistency during training. %, thereby improving irradiance prediction and overall forecast skill. 
Similarly, NVIDIA Earth-2~\cite{carpentieri2024data} integrates the FourCastNet SFNO forecasting model with dedicated radiation diagnostic modules to generate global multi-day solar irradiance forecasts. 
In industry, GraphCast~\cite{graphcast} forecasts have also been used as multi-variable meteorological inputs for power-market applications~\cite{colony2024solarcast, wang2024short}. However, these advances remain limited by coarse resolution, as they focus on model-side adaptations without leveraging satellite observations to enhance fine-grained forecasting.
% Despite these advances, the resulting irradiance products are typically provided at relatively coarse spatial resolution, which limits their suitability for fine-grained downstream forecasting. Moreover, most existing approaches primarily enhance WFM capability using model-side adaptations, while overlooking satellite observations as an auxiliary modality that could further boost high-resolution irradiance forecasting.% Motivated by this, Baguan-solar augment Baguan to deliver day-ahead solar irradiance forecasts at 0.05\textdegree resolution.
%这段结束可以再加一个逻辑：利用全球大模型的信息作为输入，提升第二阶段辐照预测是一个常见的手段，但是，这些模型缺乏xxxx。比如，缺乏更细粒度（0.05）度的预测，缺乏卫星作为辅助，无法满足下游市场对高精度细粒度数据的需求。

\subsection{Satellite-based Irradiance Forecasting}
Complementary to WFM-based irradiance products, satellite imagery provides high-frequency, high-resolution observations of cloud evolution and has become an effective auxiliary modality for day-ahead solar irradiance forecasting. 
Boussif et al. propose CrossViViT~\cite{CrossViVit}, which improves site-level irradiance prediction by incorporating geostationary satellite imagery. 
Extending to multi-site settings, Schubnel et al. develop SolarCrossFormer~\cite{schubnel2025solarcrossformer}, which couples satellite patches with station networks via graph-based cross-attention. % and introduces clear-sky irradiance as a physics-informed prior. 
For solar power forecasting, Ma et al. present FusionSF~\cite{FusionSF}, a tri-modal framework that integrates NWP outputs and satellite images, using vector quantization to align heterogeneous modalities. 
However, most multimodal methods remain limited to site-specific forecasts and lack scalable, gridded outputs for broader applications.
% Despite these advances, most multimodal approaches remain confined to site-specific forecasting, rather than producing scalable gridded irradiance fields that can directly support downstream tasks.
To bridge this gap, Bai et al. propose SolarSeer~\cite{solarseer}, an end-to-end model that uses historical satellite observations to forecast cloud cover and irradiance %over CONUS 
at 5\,km resolution, offering faster inference and lower RMSE than HRRR.
However, its reliance on satellite observations alone limits accuracy at longer lead times. %, highlighting the need to integrate WFMs with satellite data for robust high-resolution forecasting. %However, SolarSeer relies solely on satellite observations, which can limit accuracy at longer forecast horizons, motivating the integration of WFMs with satellite modalities for robust high-resolution irradiance forecasting.

\section{Multimodal Datasets}
\begin{table*}[t]
\centering
\caption{Summary of datasets used in our study. The reported spatial resolution refers to the effective resolution used in our experiments after interpolating the original products.}
\label{tab:datasets}
\begin{tabular}{lcccc}
\toprule
Dataset & Spatial Resolution & Coverage & Channels & Usage \\
\midrule
Himawari        & 0.05$^\circ$ & Asia-Pacific & Band 0.64, 3.9, 7.3, and 11.2 $\mu\mathrm{m}$ & Input during training and inference \\
CLDAS           & 0.05$^\circ$ & East Asia        & SSRD, TCDC  & Training target \\
ERA5            & 0.25$^\circ$ & Global       & U, V, T, Q, Z, TCC, SSRD, etc. & Input during training \\
Baguan forecast & 0.25$^\circ$ & Global       & U, V, T, Q, Z, TCC, SSRD, etc. & Input during inference \\
\bottomrule
\end{tabular}
\end{table*}

% \begin{table*}[t]
% \centering
% \caption{Summary of datasets used in our study.}
% \label{tab:datasets}
% \begin{tabular}{lcccc}
% \toprule
% Dataset & Spatial Resolution & Coverage & \# Channels & Usage \\
% \midrule
% Himawari        & 0.05$^\circ$ & Asia-Pacific & Band 0.64, 3.9, 7.3, and 11.2 $\mu\mathrm{m}$ & Input during training and inference \\
% CLDAS           & 0.05$^\circ$ & China        &   & Training target \\
% ERA5            & 0.25$^\circ$ & Global       & 60 & Input during training \\
% Baguan forecast & 0.25$^\circ$ & Global       & 60 & Input during inference \\
% \bottomrule
% \end{tabular}
% \end{table*}

This section describes the multimodal datasets used in our study, as summarized in Table~\ref{tab:datasets}, including geostationary satellite observations (Himawari), regional analysis fields from CLDAS, global reanalysis data (ERA5), and global WFM forecasts.  Although these datasets have different spatial coverages, all data are cropped to their common spatial intersection for subsequent analyses.

\subsection{Satellite Observations}
We utilize multi-spectral imagery from the Himawari-8/9 geostationary satellites, operated by the Japan Meteorological Agency (JMA)~\cite{Intro-to-himawari8/9}. These satellites provide full-disk observations over the Asia-Pacific region at 10-minute intervals, with spatial resolutions of 0.5–2 km depending on the channel. The visible, near-infrared, and thermal infrared bands capture critical information on cloud optical properties, aerosol loading, and atmospheric moisture. %—enabling high-temporal-resolution characterization of the initial cloud state. 
Given its low latency (<30 minutes), this data stream serves as a timely observational constraint for short-term solar irradiance prediction. %, particularly in capturing rapidly evolving cloud systems that are not resolved in coarser forecast models.
Himawari-8/9 observations from the Advanced Himawari Imager (AHI) include 13 spectral bands. The complete set of AHI bands, together with their central wavelengths and typical applications, is summarized in Table~\ref{app:tab:himawari_bands}. Following SolarSeer~\cite{solarseer}, we focus on four AHI bands: B03, B07, B10, and B14, with central wavelengths of 0.64, 3.9, 7.3, and 11.2~$\mu$m, respectively.  % As described by JMA/MSC (2024)~\cite{JMA_MSC_UtilizationMetSatData}, Band~3 (0.64~$\mu$m) measures reflected visible solar radiation and supports true-color composites as well as daytime identification of low clouds and fog. Band~7 (3.9~$\mu$m) senses emitted terrestrial radiation and includes a substantial reflected solar component during daytime; at night, it supports hotspot detection and fog/low-cloud identification via the Band~7--Band~13 brightness temperature difference. Band~10 (7.3~$\mu$m) is a water vapor channel primarily sensitive to mid-tropospheric moisture and can also respond to volcanic SO$_2$. Band~14 (11.2~$\mu$m) is a longwave infrared window channel used for cloud imaging and cloud-top characterization, and it can support surface temperature applications under clear-sky conditions.

% Geostationary provides the initial state of cloud condition, which benefits the solar irradiance prediction. 
% The Himawari-8/9 satellites, operated by the Japan Meteorological Agency (JMA), provide invaluable satellite imagery data that has revolutionized weather monitoring and analysis in the Asia-Pacific region. These geostationary
% satellites capture high-resolution imagery of East Asia, Oceania, and parts of the Pacific Ocean, offering a wealth of information for investigating atmospheric phenomena, monitoring severe weather events, and studying cloud dynamics.

\subsection{CLDAS Analysis Fields}

The CLDAS~\cite{Shi2013CLDAS} provides hourly, near-real-time land surface analysis over East Asia (0$^\circ$--65$^\circ$N, 60$^\circ$--160$^\circ$E) at an effective resolution of 0.01$^\circ$. For consistency with our model grid, we interpolate the CLDAS fields to 0.05$^\circ$ resolution. CLDAS integrates surface observations from over 30,000 automatic weather stations, FengYun satellite retrievals, radar-based precipitation estimates, and background fields from CMA’s numerical models through a statistical blending framework. We use CLDAS-derived GHI, computed from downward surface shortwave radiation (SSRD), as the target variable for model training and evaluation, where $\mathrm{GHI}\;(\mathrm{W\,m^{-2}})=\mathrm{SSRD}\;(\mathrm{J\,m^{-2}})/{3600}$. In addition, we include TCDC (total cloud cover) from CLDAS as an auxiliary predictor to provide complementary information on cloudiness conditions.

% Regional Analysis (CLDAS). The China Meteorological Administration’s Land Data Assimilation System (CLDAS) offers 0.01° resolution meteorological fields over East Asia (0–65°N, 60–160°E) with surface variables critical for regional forecasting. We employ CLDAS data (interpolate to 0.05°) from 2022/01–2024/09 for global-regional model (Mglobal−regional) training, with two independent evaluation periods defined as: Hindcast evaluation (ERA5 input): 2024/10–2024/12 and Operational evaluation (operational analysis input): 2025/01–2025/04.

\subsection{ERA5 Dataset}

ERA5~\cite{hersbach2020era5}, produced by the European Centre for Medium-Range Weather Forecasts (ECMWF), is a widely used global atmospheric reanalysis that provides comprehensive hourly estimates of a wide range of meteorological variables at a resolution of $0.25^\circ$. During the training stage, ERA5 supplies the large-scale meteorological context essential for learning the spatiotemporal dynamics of irradiance-relevant variables. In the inference stage, however, the ERA5 fields are replaced by real-time forecasts generated by Baguan~\cite{Baguan}, enabling fully operational forecasting. Additionally, while ERA5 can serve as a reference (“ground truth”) for GHI, it is inherently less accurate and coarser than higher-fidelity observational products such as CLDAS.

\subsection{Baguan Global Forecasts}
We incorporate operational forecasts from Baguan~\cite{Baguan}, a state-of-the-art data-driven global weather prediction system trained on ERA5 dataset. Baguan provides 0.25$^\circ$-resolution forecasts of key atmospheric variables, including GHI, at hourly lead times up to 14 days. %The system is initialized from global reanalysis fields and trained on historical ERA5 data. 
Within our framework, during inference, Baguan supplies the large-scale meteorological context and a coarse predictive signal for GHI, which we further refine using high-resolution satellite observations. In addition, Baguan serves as a WFM baseline, as it directly produces GHI forecasts at 0.25$^\circ$ resolution.

\begin{figure*}[t!]
    \centering
    \includegraphics[width=1.0\textwidth]{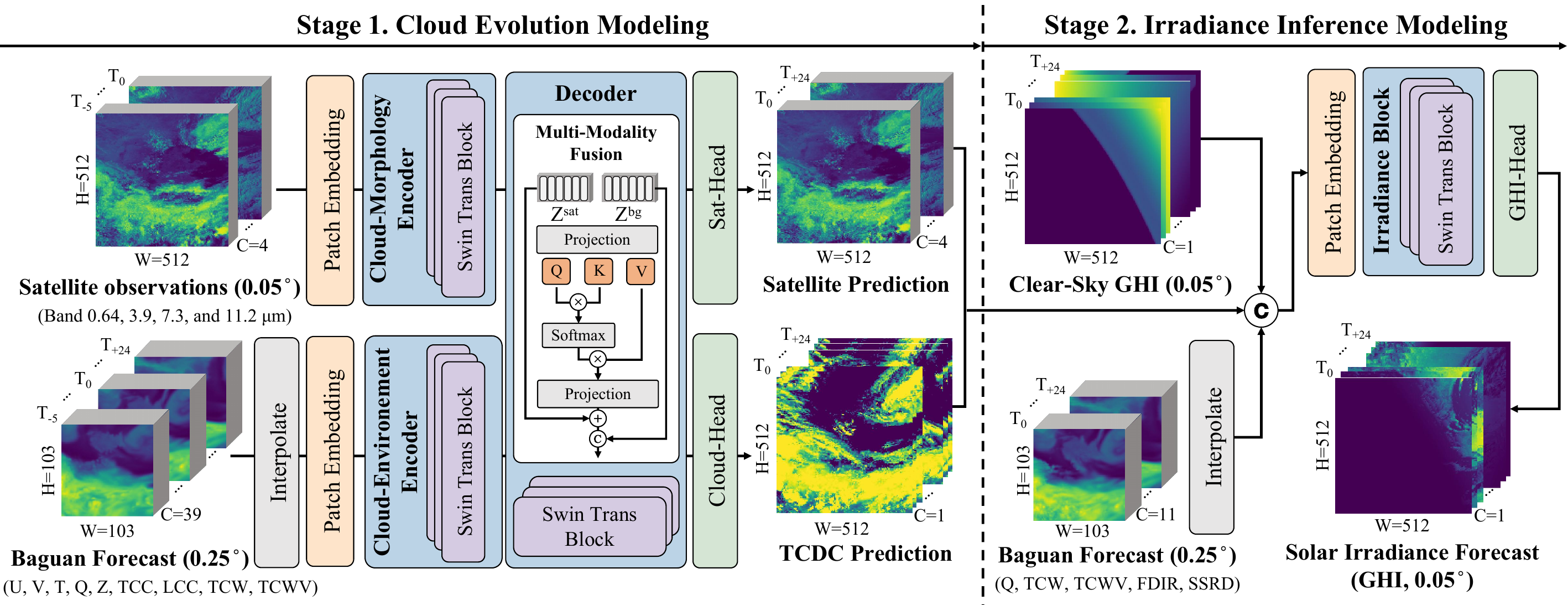}
    \caption{Baguan-solar model architecture. Baguan-solar uses a two-stage Swin Transformer framework that fuses Himawari satellite observations with Baguan forecasts to first predict cloud-related intermediates (satellite fields and TCDC) and then infer 24~h high-resolution GHI.}
    \label{fig:model_architcture}
\end{figure*}

\section{Baguan-solar Framework}

% \subsection{Model Architecture.}
As illustrated in Figure~\ref{fig:model_architcture}, Baguan-solar contains two stages: (i) \emph{cloud evolution modeling} and (ii) \emph{irradiance inference modeling}.
Stage~1 explicitly forecasts the future 24~h cloud cover and satellite images by fusing historical 6~h multi-spectral geostationary satellite observations and 30~h Baguan weather forecasts (spanning both the past 6~h and the subsequent 24~h).
Stage~2 then infers the future 24~h GHI forecast by combining Stage~1 cloud-aware outputs with clear-sky GHI and radiation-relevant Baguan variables. Specifically, clear-sky GHI is estimated by the Ineichen--Perez model and is a deterministic function of longitude, latitude, and time (see Appendix~\ref{app:sec:fast_ghi}). 
Both stages are implemented with Swin Transformer~\cite{liu2021swintransformerhierarchicalvision} backbones to capture multi-scale spatial structures efficiently while preserving high-resolution outputs via patch embedding and patch recovery.

\subsection{Stage 1: Cloud Evolution Modeling.}
Accurate GHI forecasting critically depends on predicting cloud evolution, since clouds dominate radiative attenuation and introduce strong spatiotemporal nonlinearity. 
Thus, Stage~1 is formulated as an explicit cloud-field forecasting task.
Specifically, it leverages two complementary inputs: 
(i) historical satellite observations $\X^{\rm sat}_{t-5:t} \in \mathbb{R}^{6\times 4 \times H_{\rm sat} \times W_{\rm sat}}$, where the four spectral channels are defined in Section~3.1; 
and (ii) Baguan weather forecasts $\X^{\rm bg}_{t-5:t+24} \in \mathbb{R}^{30\times C_1 \times H_{\rm bg} \times W_{\rm bg}}$. 
We select a total of $C_1=39$ Baguan channels, including moisture, cloud state, thermodynamic, and dynamical conditions (see Appendix Table~\ref{tab:dataset_baguan}). 
All Baguan variables are interpolated to the satellite grid.

Architecturally, Stage~1 uses two Swin Transformer encoders to disentangle cloud morphology and atmospheric forcing.
We first project satellite observations and Baguan forecast fields into patch tokens via modality-specific patch embeddings $\phi_{\rm sat}(\cdot)$ and $\phi_{\rm bg}(\cdot)$, and then feed them into a \emph{Cloud-Morphology Encoder} and a \emph{Cloud-Environment Encoder}, respectively.
Each encoder consists of \(8\) stacked Swin Transformer blocks with residual connections, layer normalization and multi~head attention. We use a patch size of \(8\times 8\), a window size of \(16\), an embedding dimension of \(256\), and \(2\) attention heads in each transformer layer.
The Cloud-Morphology Encoder extracts multi-scale cloud textures and boundary cues from satellite imagery to produce $\Z^{\rm sat}$, while the Cloud-Environment Encoder encodes the interpolated Baguan variables to capture dynamical and thermodynamical conditions, yielding $\Z^{\rm bg}$.
we then apply cross-attention to inject environmental guidance into the satellite representation, and concatenate the enhanced satellite tokens with the Baguan tokens to form a fused representation $\Z^{\rm fused}$:
\begin{equation}
    \Z^{\rm sat}=\mathrm{Enc_{\rm sat}}({\phi}_{\rm sat}(\X^{\rm sat})),\quad \Z^{\rm bg}=\mathrm{Enc_{\rm bg}}({\phi}_{\rm bg}(\X^{\rm bg})), 
\end{equation}
\begin{equation}
    \Z^{\rm fused} = \mathrm{cat}(\Z^{\rm sat} + \mathrm{Attn}(\Z^{\rm sat}W_Q, \Z^{\rm bg}W_K, \Z^{\rm bg}W_V), \Z^{\rm bg}).
\end{equation}
Finally, $\Z^{\rm fused}$ is processed by a shared Swin Transformer decoder followed by two task-specific heads to jointly generate the day-ahead total cloud cover forecast $\hat \Y^{\rm TCDC}_{t+1:t+T_{\rm out}}$ and future satellite predictions $\hat \Y^{\rm sat}_{t+1:t+T_{\rm out}}$.
\begin{equation}
    [\hat \Y^{\rm TCDC}_{t+1:t+T_{\rm out}}, \hat \Y^{\rm sat}_{t+1:t+T_{\rm out}}] = \mathrm{Dec}(\Z^{\rm fused}).
\end{equation}

\subsection{Stage 2: Irradiance Inference Modeling.}
Stage~2 infers irradiance by integrating the cloud-aware intermediate outputs from Stage~1 with additional meteorological and physical priors.
Specifically, we construct the Stage~2 input by concatenating the Stage~1 predictions $\hat{\Y}^{TCDC}_{t+1:t+24}$ and $\hat{\Y}^{\rm sat}_{t+1:t+24}$, together with the clear-sky GHI \(\X^{\rm clear-sky}_{t+1:t+24}\) and Baguan forecast variables $\X^{\rm bg}_{t+1:t+24} \in \mathbb{R}^{T_{\rm out}\times C_2 \times H_s \times W_s}$ with $C_2=11$ radiation-relevant channels.
The concatenated tensor is patch-embedded and processed by a Swin Transformer backbone with $8$ stacked Swin blocks, patch size $P{=}8$, and hidden dimension $D{=}256$. 
A Solar Irradiance head then performs patch recovery to restore the original spatial resolution and outputs multi-step GHI forecasts:
\begin{equation}
    \hat{\Y}^{ghi} = \mathrm{Irr}(\mathrm{cat}(\X^{\rm clear-sky}, \X^{\rm bg}, \hat{\Y}^{\rm TCDC}, t{\Y}^{\rm sat})).
\end{equation}
We train the two-stage model in an end-to-end manner with a weighted multi-task objective over three prediction targets. We use mean squared error (MSE) as the loss function for all tasks:
\begin{equation}
\mathcal{L} = \lambda_{\mathrm{sat}} \mathcal{L}_{\mathrm{sat}}
+ \lambda_{\mathrm{TCDC}} \mathcal{L}_{\mathrm{TCDC}}
+ \lambda_{\mathrm{ghi}} \mathcal{L}_{\mathrm{ghi}},
\end{equation}
where $\lambda_{\mathrm{sat}}=1$, $\lambda_{\mathrm{TCDC}}=0.5$, and $\lambda_{\mathrm{ghi}}=1$ in all experiments.

\subsection{Implementation \& Evaluation}

The multimodal dataset from 2022--2024 is used for model development and is split into training and validation subsets at a 0.9:0.1 ratio. Data from 2025 are reserved exclusively for testing to provide an independent evaluation. %The model is first trained on ERA5 data and subsequently fine-tuned using Baguan prediction data. 
All inputs are cropped to a $512 \times 512$ pixel domain covering East Asia. Baguan-solar is trained on 8 NVIDIA A100 GPUs with a batch size of 4. The training of Baguan-solar uses the scheduler-free optimizer~\cite{schedulerFree}, which removes the need for an explicit learning-rate schedule while maintaining stable convergence. %accompanied by a weight decay parameter set to 0.05.
The complete set of model hyperparameters is provided in Appendix~\ref{app:sec:hyperparams}.

% \subsection{ Metrics}
To assess the quality of GHI forecasts, we adopt the root mean squared error (RMSE) as the primary metric, consistent with prior studies~\cite{solarseer,rasp2024weatherbench2benchmarkgeneration,Baguan}. RMSE is computed as:
\begin{equation}
\mathrm{RMSE}=\sqrt{\frac{1}{n}\sum_{i=1}^{n}\left(y_i-\hat{y}_i\right)^2},
\label{eq:rmse}
\end{equation}
where $n$ denotes the number of samples in the test set, and $y_i$ and $\hat{y}_i$ are the observed and predicted GHI values. Smaller RMSE values correspond to more accurate forecasts.

% \begin{equation}
% \mathrm{MAE}=\frac{1}{n}\sum_{i=1}^{n}\left|y_i-\hat{y}_i\right|,
% \label{eq:mae}
% \end{equation}

% We further define a first-order difference metric $\mathrm{Diff}$ to characterize the fluctuation of solar irradiance and cloud cover:
% \begin{equation}
% \mathrm{Diff}=y_t-y_{t-1},
% \label{eq:diff}
% \end{equation}
% where $y_t$ and $y_{t-1}$ denote the solar irradiance (or total cloud cover) at the current and previous time steps, respectively. Based on the predicted values, we can compute the corresponding predicted first-order differences, whose performance is then evaluated using MAE and RMSE.

\section{Experiments}

\subsection{Benchmarking on CLDAS}

\subsubsection{Baselines and Experimental Setup}

\begin{table*}[h]
\caption{Benchmark comparison (RMSE) for solar irradiance forecasting (GHI $\mathrm{W\,m^{-2}}$) over East Asia in 2025, evaluated for forecasts initialized at 00:00 and 12:00 UTC. Results are reported at lead times of 1~h, 2~h, 3~h, 6~h, 12~h, and 24~h, as well as the average RMSE over 1--24~h (Avg.).}
\label{Tab1:benchmark}
\begin{center}

\begin{tabular}{lrrrrrrrrr}
\toprule
\multirow{2}{*}{Model} & \multirow{2}{*}{Type} & \multirow{2}{*}{Inputs} & \multicolumn{7}{c}{RMSE $(\mathrm{W\,m^{-2}}$) } \\
\cline{4-10}
& & & Avg. & 1~h & 2~h & 3~h & 6~h & 12~h & 24~h \\
\midrule
Mean & Statistical & None & 83.89 & -- & -- & -- & -- & -- & -- \\
Clear-sky & Statistical & None & 113.54 & -- & -- & -- & -- & -- & --  \\
% \hline
Baguan~\cite{Baguan} & weather foundation model & Gridded initial field &  58.17 & 49.53 & 58.57 & 65.50 & 77.98 & 37.86 & 37.13 \\
EC IFS     & NWP & Gridded initial field & 54.46 & 47.48 & 59.86 & 70.65 & 72.23 & 37.59 & 36.50 \\
% \hline
SolarSeer~\cite{solarseer}       & Extrapolation-based & Satellite & 53.09 & 36.07 & 51.40 & 64.40 & 68.75 & 33.14 & 35.47 \\
Two-stage Unet           & Extrapolation-based & Satellite &  59.10 & 47.03 & 60.32 & 72.15 & 74.79 & 39.10 & 41.78  \\
Two-stage Swin  & Extrapolation-based & Satellite & 49.89 & 32.74 & 46.77 & 59.23 & 65.33 & 31.46 & 33.51 \\

\midrule
Baguan-solar   & Multimodal & ERA5 \& Satellite &  41.21 & 29.99 & 43.52 & 55.04 & 57.66 & 24.80 & 24.20  \\
\textbf{Baguan-solar (oper.)}  & Multimodal  & Baguan forecasts \& Satellite & 41.87 & 30.31 & 43.64 & 55.34 & 57.98 & 25.18 & 25.04 \\
\bottomrule
\end{tabular}

\end{center}
\end{table*}

% \begin{table*}[h]
% \caption{Benchmark RMSE($\downarrow$) comparison for solar irradiance forecasting over China in 2025. at the initial times of UTC 00:00 and 12:00, with lead times of 1h, 2h, 3h, 6h, 12h, and 24h and the mean of 1 to 24h.}
% \label{Tab1:benchmark}
% \begin{center}

% \begin{tabular}{lrrrrrrr}
% \hline
% \multirow{2}{*}{Model names} & \multirow{2}{*}{Model types} & \multirow{2}{*}{Model types} & \multicolumn{7}{c}{RMSE} \\
% \cline{2-8}
%  & mean (24h) & 1h & 2h & 3h & 6h & 12h & 24h \\
% \hline
% Mean & 83.89 & - & - & - & - & - & - \\
% Clear Sky & xx & - & - & - & - & - & -  \\
% \hline
% Baguan predict &  58.17 & 49.53 & 58.57 & 65.50 & 77.98 & 37.86 & 37.13 \\
% EC predict     & 54.46 & 47.48 & 59.86 & 70.65 & 72.23 & 37.59 & 36.50 \\
% \hline
% Unet           &  59.10 & 47.03 & 60.32 & 72.15 & 74.79 & 39.10 & 41.78  \\

% SolarSeer      & 53.09 & 36.07 & 51.40 & 64.40 & 68.75 & 33.14 & 35.47 \\

% 2Swin(2stage) & 49.89 & 32.74 & 46.77 & 59.23 & 65.33 & 31.46 & 33.51 \\
% % 2Swin by Xinyue &  &  &  &  &  &  &  \\
% 2Swin + era5    &  \textbf{41.21} & 29.99 & 43.52 & 55.04 & 57.66 & 24.80 & 24.20  \\
% 2Swin + baguan predict  & 41.87 & 30.31 & 43.64 & 55.34 & 57.98 & 25.18 & 25.04 \\
% \hline
% \end{tabular}

% \end{center}
% \end{table*}

We evaluate Baguan-solar against the following established benchmarks:

\textbf{Operational Weather Models:}
\begin{itemize}
    \item \textbf{Baguan}~\cite{Baguan}: An operational weather foundation model designed for renewable energy, providing irradiance-relevant parameters at $0.25^\circ$ resolution.
    \item \textbf{EC IFS}: ECMWF (EC) Integrated Forecasting System (IFS) is a high-resolution ($0.1^\circ$) NWP system that directly outputs surface solar radiation downward (SSRD), serving as a robust benchmark.
\end{itemize}

\textbf{Satellite-based Models:}
\begin{itemize}
    \item \textbf{SolarSeer}~\cite{solarseer}: A state-of-the-art, satellite-based nowcasting model, retrained our dataset for a region-fair evaluation.
    \item \textbf{Two-stage U-Net}: Our two-stage U-Net baseline, built following SolarSeer’s two-stage design.
    \item \textbf{Two-stage Swin}: Our two-stage Swin Transformer baseline, also built following SolarSeer’s two-stage design.
\end{itemize}

\textbf{Statistical Baselines:}
\begin{itemize}
    \item \textbf{Mean}: Predicts the historical average GHI for each hour from the training period (2022--2024).
    \item \textbf{Clear-sky}: Estimates the theoretical GHI under cloud-free conditions based on spatiotemporal coordinates.
\end{itemize}

Both weather model forecasts are bilinearly interpolated to $0.05^\circ$ for a consistent comparison. All models are evaluated on a common test set comprising data from the year 2025.

\subsubsection{Overall Performance Comparison}
% In this study, Baguan-solar is benchmarked against a range of baseline and state-of-the-art models as shown in Table~\ref{Tab1:benchmark}. it is observed that Baguan-solar (oper.) outperforms the best baseline (Two-stage Swin) by reducing xx\% in RMSE. and Baguan-solar outperforms original Baguan (0.25 degree) by reducing xx\% in RMSE. Among all the baseline, Extrapolation-based models (SolarSeer \& Two-stage Swin) show advantage due to their training target is 0.05 degree CLDAS data. they can easily outperform Baguan and EC at 0.25 degree.

We evaluate Baguan-solar against a comprehensive set of baselines.
As summarized in Table~\ref{Tab1:benchmark}, the operational Baguan-solar achieves the best performance, reducing the average RMSE by  16.08\% compared to the strongest baseline, the Two-stage Swin, and reducing RMSE by 28.02\% relative to the Baguan forecasts. 
The extrapolation-based methods, especially SolarSeer and Two-stage Swin, show clear advantages over operational weather models. This performance gap stems primarily from their training on high-resolution ($0.05^\circ$) CLDAS data. 
The finer spatial resolution allows these models to better capture local GHI variability, leading to systematic gains over coarser-resolution approaches such as Baguan at $0.25^\circ$. 
We further evaluate two variants of Baguan-solar. 
The idealized variant uses ERA5 reanalysis as input, which assumes perfect knowledge of future atmospheric conditions and is not operationally feasible. 
The operational variant instead uses Baguan forecasts as input, thereby emulating real-time deployment through reforecast experiments.
The results show that the idealized version slightly outperforms the operational version, but the gap is small, suggesting that Baguan’s short-term weather forecasts are highly accurate and closely approximate actual atmospheric conditions and introduce only limited degradation in downstream GHI prediction.

% We compare Baguan-solar with statistic baselines (Mean and Clear sky), AI weather models (Baguan), NWP (EC), and several extrapolation-based model (SolarSeer, Unet, Two-stage Swin) as shown in Table~\ref{Tab1:benchmark}. We present two versions of Baguan-solar, first uses ERA5 as input which is an ideal condition of future weather state and not applicable in practical use. Second is the operational version and use Baguan forecasts as input. This result is we mimic the online situation of perform reforecasts. We oberseve that Baguan-solar in ideal condition is better than Baguan operational but the differece is small, which indicates that the prediction of Baguan for short-term weather condition is so close to the real data so that influence very little to final results of Baguan-solar.

\subsubsection{Lead-time-dependent Forecast Skill}
% 画图,同solarSeer,24小时的曲线。

\begin{figure}[t!]
    \centering
    \includegraphics[width=0.48\textwidth]{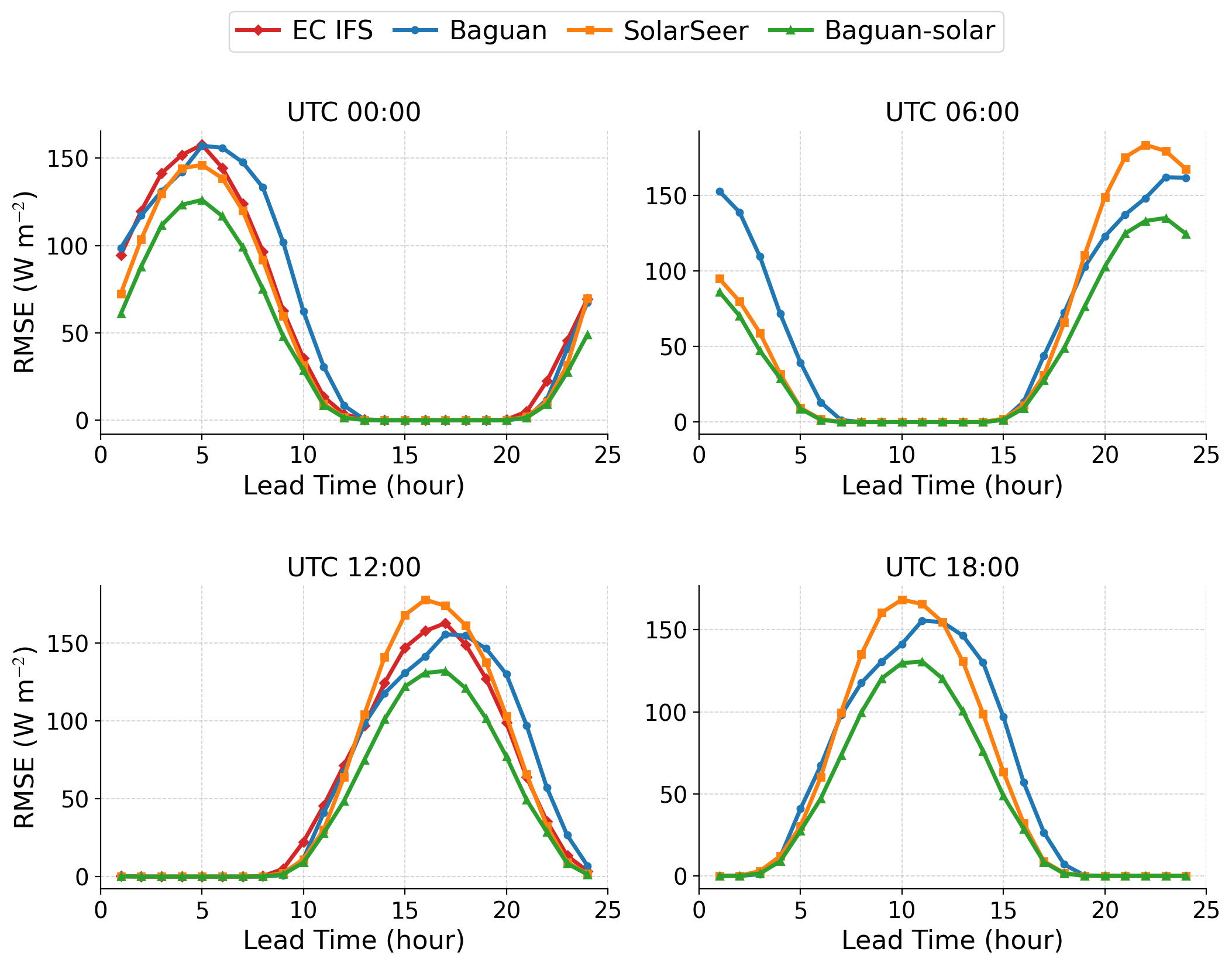}
    \caption{RMSE of GHI forecasts as a function of lead time (1–-24~h) for four initialization times (UTC 00:00, 06:00, 12:00, and 18:00), evaluated over a 512 × 512 gridded domain.}
    \label{fig:rmse_curve}
    \vspace{-1.5em}
\end{figure}

\begin{figure*}[t!]
    \centering
    \includegraphics[width=1.0\textwidth]{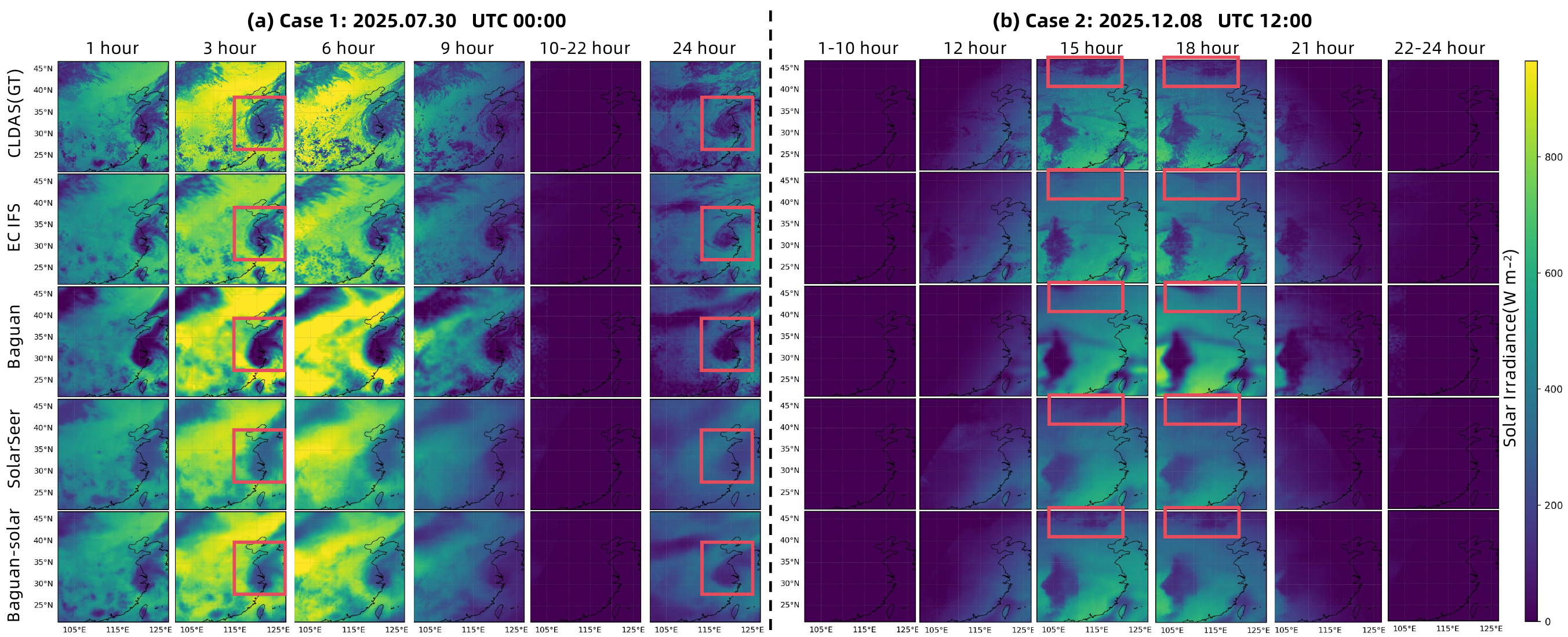}
    \caption{Qualitative comparison of GHI forecast fields for two representative cases initialized at UTC 00:00 and 12:00.}
    \label{fig:case_vis}
\end{figure*}

Figure~\ref{fig:rmse_curve} shows how the RMSE of GHI forecasts varies with lead time from 1 to 24~h, comparing Baguan-solar with EC IFS, Baguan~\cite{Baguan}, and SolarSeer~\cite{solarseer}. Across all initialization times (UTC 00:00, 06:00, 12:00, and 18:00), Baguan-solar consistently achieves the lowest errors at every lead time. %, indicating robust improvements in both short- and day-ahead forecasting skill. 
In addition, a pronounced diurnal cycle is observed in the error curves: errors increase during local daytime, peaking around midday when GHI magnitude is highest, and decrease toward nighttime. During night hours, when GHI is effectively zero, RMSE approaches zero for all methods, reflecting the negligible forecasting uncertainty under no-sun conditions. We also observe that SolarSeer, as an extrapolation-based model, degrades markedly with increasing lead time, consistent with the accumulation of cloud-motion errors and the lack of large-scale dynamical constraints.

\subsection{Ablation Studies}
\begin{table}[t]
\centering
\caption{Ablation studies on modality contributions and stage-wise decoupling in Baguan-solar. "Only S1" means forecasting GHI and TCDC in a single stage, and then using Clear-sky GHI for post-processing to mask out the night. Results are reported at lead times of 1~h, 2~h, 3~h, 6~h, 12~h, and 24~h, as well as the average RMSE over 1--24~h (Avg.).}
\label{Tab2:ablation}
\setlength{\tabcolsep}{3pt}
\renewcommand{\arraystretch}{1.15}
\small

\begin{tabularx}{\columnwidth}{%
    l
    rrrrrrr
}
\toprule
\multirow{2}{*}{Exp. (setting)} &
\multicolumn{7}{c}{RMSE ($\mathrm{W\,m^{-2}}$)} \\
\cmidrule(lr){2-8}
& Avg. & 1~h & 2~h & 3~h & 6~h & 12~h & 24~h \\
\midrule
\textbf{Baguan-solar (S1+S2)}
& 41.87 & 30.31 & 43.64 & 55.34 & 57.98 & 25.18 & 25.04 \\
w/o Baguan
& 49.89 & 32.74 & 46.77 & 59.23 & 65.33 & 31.46 & 33.51 \\
w/o satellite
& 42.66 & 37.35 & 49.54 & 59.36 & 59.03 & 25.05 & 24.60 \\
\midrule
Baguan-solar (Only S1)
& 45.50 & 32.34 & 44.86 & 56.38 & 59.73 & 35.86 & 28.92 \\
w/o TCDC
& 48.30 & 35.21 & 47.30 & 58.37 & 61.87 & 41.00 & 32.62 \\

\bottomrule
\end{tabularx}
\end{table}

In this section, we compare variants of Baguan-solar to quantify the modality contributions and stage-wise decoupling. 
As shown in Table~\ref{Tab2:ablation}, our ablation studies reveal several key insights. 
Removing Baguan forecasts significantly increases the average RMSE by 19.15\%. %, highlighting the significance of incorporating Baguan forecast. 
The disparity persists across all lead times, especially at longer lead times, with RMSE increasing by 25.18\% at 12~h and 33.82\% at 24~h, where the satellite-only extrapolation struggles to capture cloud formation or dissipation. 
By contrast, Baguan forecasts provide thermodynamic and dynamical conditions that better constrain the evolution of cloud fields, leading to accurate GHI forecasts at longer lead times.
Removing satellite increases the average RMSE by 1.88\%, while the short-term error increase substantially by 23.22\% at 1~h and 13.5\% at 2~h . This result emphasizes the role of satellite imagery in capturing fine-scale cloud morphology and boundary motion.
In addition, simplifying the two-stage framework to single-stage increases the average RMSE by 8.67\%, and further removing TCDC supervision increases it by 15.36\%. 
These results indicate that decoupling cloud evolution provide a physically grounded intermediate constraint that makes GHI variations attributable to forecast cloud occurrence and motion, thereby improving physical consistency.

% \subsection{Scaling law}

\subsection{Qualitative Results}

We present two representative cases for qualitative evaluation. 
The first case (on 2025.07.30) features an organized vortex over East Asia, forming a distinct pattern that low-GHI core surrounded by higher GHI.
The Second case (on 2025.12.08), for a typical winter day, features a distinct low-GHI belt over North East Asia.
Across both cases, EC IFS follows the overall structure reasonably well but tends to under-suppress the low-GHI core with narrow range and sharp boundaries.
Baguan exhibits a systematic bright bias in both cases. 
SolarSeer is consistently over-smoothed, blurring cloud-band boundaries and gradients. In particular, it fails to retain the vortex-related signature at 24~h in the first case. 
In contrast, Baguan-solar provides the most balanced reconstruction in both morphology and amplitude. 
It captures fine-scale structures and transitions for short-term forecast. Although performance degrades with increasing lead time, it still retains the reasonable intensity and spatial extent compared to the other methods.

\vspace{-0.6em}

\subsection{Modality Importance Analysis}
To justify our two-stage design and the inclusion of multimodal data, we use Integrated Gradients (IG)~\cite{Sundararajanetal2017} to quantify how much each input modality, satellite versus ERA5 (or Baguan forecasts), contributes to RMSE reduction across lead times 1--24~h. The analysis is performed over 2400 samples and 24 different initialization hours, ensuring robustness to diurnal cycles and variability across initialization times.
\begin{figure}[t!]
    \centering
    \includegraphics[width=0.39\textwidth]{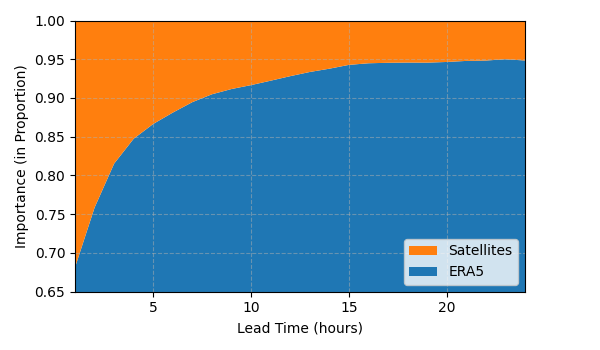}
    \vspace{-1.2em}
    \caption{Importance of ERA5 vs Satellites across lead times.}
    \label{fig:ig}
    \vspace{-1.2em}
\end{figure}
The feature importance in Figure~\ref{fig:ig} reveals a clear temporal dynamics: the satellite input contributes significantly (10-31\%) to RMSE at short lead times (1--6~h), but it rapidly drops after 6~h and is ultimately below 5\% by 24~h. In contrast, ERA5's contribution starts high (68.1\%) and steadily increases to over 95\% at 24~h.

This pattern justifies two key aspects of our design. First, it highlights the need for including ERA5 (or Baguan forecasts) during training. Although high-resolution satellite data can extrapolate well and shine in the nowcasting of GHI, 
%%%capture short-term cloud structures and dominate short-term GHI variability, 
meteorological fields provide essential large-scale dynamical and thermodynamical information to achieve skillful day-ahead forecasts. The finding also coincides with the ablation study in Table~\ref{Tab2:ablation} and the GHI forecast results in Figure~\ref{fig:rmse_curve}.
Second, it validates the rationale for the  architecture that decouples the forecasting problem into two physically grounded subtasks. Stage 1 is designed to handle the time-varying subtask by learning to dynamically weigh satellite and meteorological inputs according to the forecast horizon, avoiding the degeneration of pure extrapolation. This specialization allows Stage 1 to model difficult cloud evolution under shifting modality dominance as a standalone task. Stage 2 then focuses on the more stable transformation from predicted clouds to GHI, relying only on physical priors such as clear-sky GHI. By separating these concerns, our architecture reduces overall learning difficulty and improves forecast skill across all lead times.

\section{Deployment}
\subsection{Operational Deployment in East China}

Since July 2025, Baguan-solar has been deployed online to support operational solar power forecasting in an eastern province in China, which has the highest solar power capacity of 
918.4~GW among all provinces. It is co-deployed with the Baguan weather forecasting system and shares a similar operational pipeline: Baguan is executed four times per day (UTC 00:00, 06:00, 12:00, and 18:00) 
%%%using global initial conditions from the ECMWF (EC) analysis 
and produces weather forecasts with lead times up to 14 days. 
%%Due to the extensive acquisition, quality control, and assimilation required to generate the EC analysis (from in situ, aircraft/radiosonde, and satellite observations), the EC initial fields typically become available to Baguan about 6 hours after the analysis time; 
Each time, it takes 0.5~h for Baguan to perform inference on two GPUs. 
On the other hand, the four AHI bands (B03, B07, B10, and B14) imagery data from Himawari-8/9 geostationary satellites are collected every 10 minutes. 
%%%Baguan then needs roughly 0.5 hour to complete the rolling forecast on two GPUs. 
Building on the latest available Baguan outputs and satellite data, Baguan-solar runs at a higher frequency (hourly) to provide 24~h high-resolution GHI forecasts,
%%%while satellite observations can be accessed with much lower latency (about 30 minutes), 
offering a faster but less dynamically constrained view of the evolving atmosphere. The weather forecasts, including the GHI forecasts, are provided to the downstream applicaitons, such as solar power forecasting model and electric load forecasting models~\cite{zhu2023eforecaster}. 

%%As this system is deployed for decision-making, high data reliability is strictly required. In the production system, we actually import multiple data sources to ensure 

% In terms of time latency, the Baguan model depends on the initial conditions provided by the ECMWF (EC) analysis. To generate these EC initial fields, a substantial data acquisition and assimilation process is required, including the collection and quality control of conventional in situ observations, aircraft and radiosonde measurements, and various satellite data products. As a consequence, there is typically a delay of about 6 hours between the analysis time and the availability of the corresponding EC initial field to Baguan. Once the EC initial field is received, Baguan itself requires on the order of 1 hour to complete the rolling forecast and produce future-state estimates. By contrast, satellite observations can in principle be accessed and exploited with a much shorter latency, on the order of 30 minutes, offering a more rapid—but less dynamically constrained—view of the evolving system. 

\subsection{Operational Verification with Pyranometer Sites}

We collect GHI measurements from 246 sites equipped with pyranometers in this province and use these in-situ observations to verify Baguan-solar forecasts. Although the previous section uses the CLDAS reanalysis data as the reference due to its fine spatial resolution, pyranometers in these sites provide a more direct and accurate measure of surface GHI and therefore constitute a stricter benchmark for operational validation. To further assess which gridded product better matches the observations, we compute the averaged RMSE between the site measurements and the ERA5/CLDAS fields interpolated to the station locations. %we compute the RMSE of gridded ERA5/CLDAS fields and site measurements. 
ERA5 yields a higher RMSE (77.85) than CLDAS (66.69), indicating that CLDAS provides a more reliable gridded reference over our study region; this also supports our choice of CLDAS as the ground truth for model training and evaluation. 

\begin{figure}[t]
    \centering
    \includegraphics[width=0.48\textwidth]{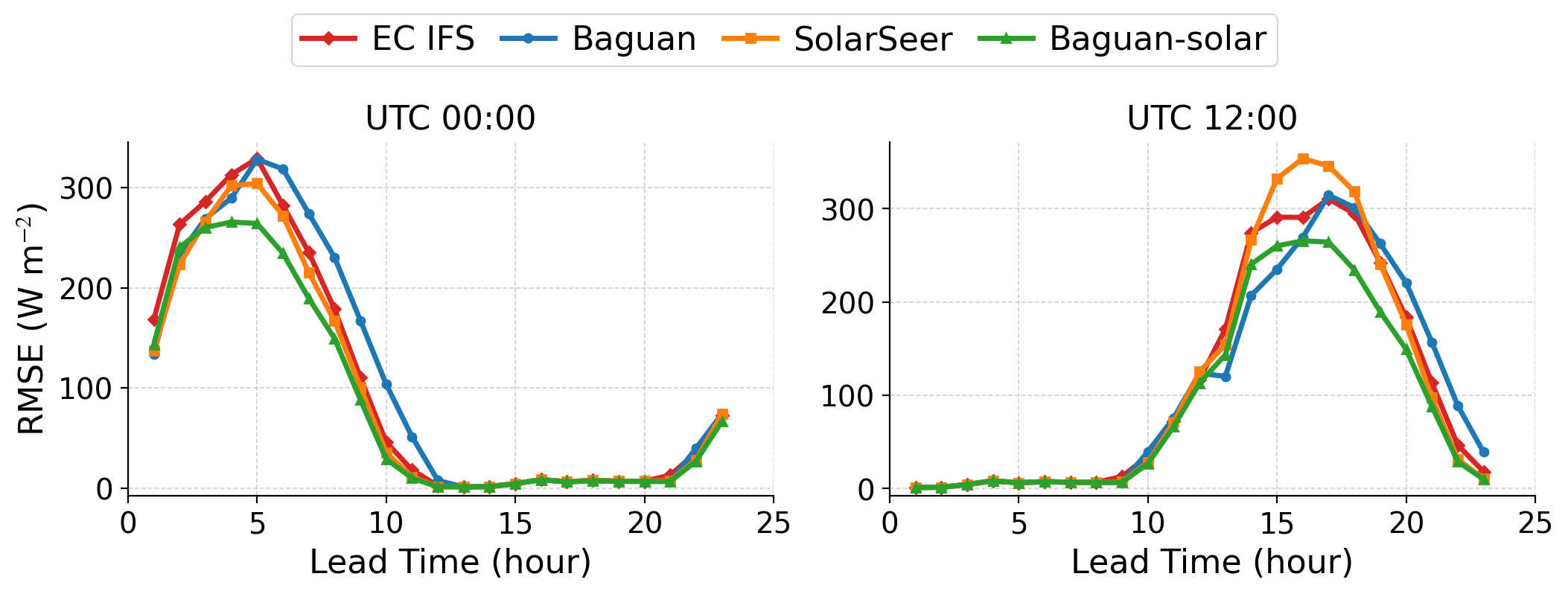}
    \includegraphics[width=0.47\textwidth]{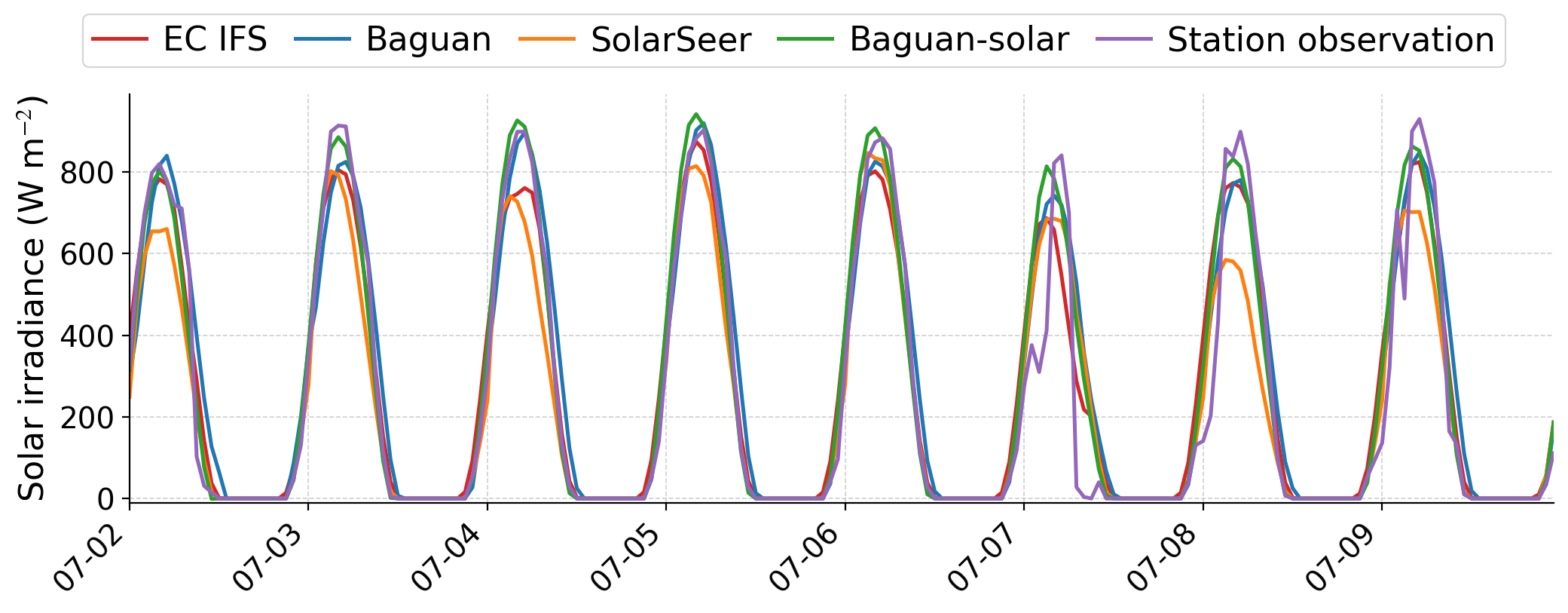}
    \caption{(Top) RMSE of GHI forecasts as a function of lead time (1--24~h) for two initialization times (UTC 00:00 and 12:00), averaged over 246 sites.
(Bottom) Visualization of GHI forecasts for one photovoltaic site over a one-week period.}
    \label{fig:rmse_curve_station}
    \vspace{-1.6em}
\end{figure}

Figure~\ref{fig:rmse_curve_station}~(top) illustrates the RMSE of GHI forecasts for Baguan-solar and other baselines. It shows that Baguan-solar consistently achieves the lowest average errors across all initialization times. SolarSeer, as an extrapolation-based model, degrades remarkably with increasing lead time, with the most pronounced deterioration in the 12--24~h range for the UTC 12:00 initialization. EC IFS and Baguan exhibit similar overall performance; however, Baguan tends to perform worse in the afternoon. In the one-week case study at the photovoltaic site in July 2025 (Figure~\ref{fig:rmse_curve_station}~(bottom)), Baguan-solar shows closer agreement with the site observations than the baseline forecasts, capturing the peak irradiance and the overall temporal variability more accurately.

\section{Conclusion and Future Work}

We presented Baguan-solar, a two-stage model that fuses weather foundation model forecasts and satellite imagery to deliver accurate, fine-grained ($0.05^\circ$) day-ahead solar irradiance predictions. In evaluations, our model surpasses strong baselines (EC IFS, Baguan, and SolarSeer) not only in overall accuracy but also in tracking rapid irradiance changes caused by clouds. It has been in operational use in China since 2025.

Future work includes extending the system globally using multi-satellite data, incorporating uncertainty quantification via ensemble methods, and exploring end-to-end co-training with upstream weather foundation models for improved physical consistency.

\clearpage


%\newpage
\bibliographystyle{ACM-Reference-Format}
\bibliography{sample-base,iclr2025_conference, related_work}

% \newpage
% \onecolumn % Start one-column mode here

\appendix
\section{Appendix}
\newcolumntype{M}[1]{>{\centering\arraybackslash}m{#1}}

\begin{table*}[h]
\caption{Comparison of representative weather foundation models and solar irradiance forecasting methods, including their inputs/outputs, resolution, region, and whether irradiance-related variables are explicitly modeled.}
\label{app:tab:compare_wfms}
\begin{center}
% \begin{tiny}
% \usepackage{multirow}
% \resizebox{1\textwidth}{!}

\begin{tabular}{M{2cm}M{3cm}M{3cm}M{1.8cm}M{1.5cm}M{2.2cm}M{1.8cm}}
\toprule
Model & Input Variable & Output Variable & \makecell{Spatial\\Resolution} & Region  & Accuracy & \makecell{Irradiance-\\related Variable} \\
\midrule
Graphcast~\cite{graphcast} & {U10, V10, T2M, MSLP, TP, U, V, Q, Z, T, W} & {U10, V10, T2M, MSLP, TP, U, V, Q, Z, T, W} & 0.25\textdegree & Global & beats EC IFS  & $\times$ \\
Pangu-Weather~\cite{pangu_nature} & {U10, V10, T2M, MSLP, U, V, Q, Z, T} & {U10, V10, T2M, MSLP, U, V, Q, Z, T} & 0.25\textdegree & Global & beats EC IFS  & $\times$ \\
Fengwu~\cite{chen2023fengwu} & {U10, V10, T2M, MSLP, U, V, Q, Z, T} & {U10, V10, T2M, MSLP, U, V, Q, Z, T} & 0.25\textdegree & Global & beats EC IFS  & $\times$ \\
Fuxi~\cite{Fuxi_nature} & {U10, V10, T2M, MSLP, TP, U, V, Q, Z, T} & {U10, V10, T2M, MSLP, TP, U, V, Q, Z, T} & 0.25\textdegree & Global & beats EC IFS  & $\times$ \\
\midrule
Baguan & {U10, U100, V10, V100, T2M, MSLP, TP, TCC, LCC, FDIR, SSRD, TCW, TCWV, TP, SP, U, V, Q, Z, T} & {U10, U100, V10, V100, T2M, MSLP, TP, TCC, LCC, TCW, TCWV, TP, SP, \textbf{FDIR}, \textbf{SSRD}, U, V, Q, Z, T} & 0.25\textdegree & Global & beats EC IFS  & $\checkmark$ \\
% Fuxi-2.0 with SSRD
% SunCastNet: ground truth 是向日葵反演得到的。
\midrule
SolarSeer~\cite{solarseer} & Satellite & \textbf{SSRD} & \textbf{0.05}\textdegree & the CONUS & beats HRRR & $\checkmark$\\
\midrule
\textbf{Baguan-solar} & Satellite $\And$ Baguan forecasts & \textbf{SSRD} & \textbf{0.05}\textdegree & China & beats all baselines
 & $\checkmark$\\
\bottomrule
\end{tabular}

% \end{tiny}
\end{center}
\end{table*}

\begin{table*}[htbp]
\centering
\caption{Himawari-8/9 (AHI) spectral bands and typical applications.}
\label{app:tab:himawari_bands}
\begin{tabular}{clll}
\hline
Band & Center wavelength ($\mu$m) & Type & Typical applications \\
\hline
B01 & 0.47 & Visible & Aerosol/land--ocean contrast; thin cloud (daytime) \\
B02 & 0.51 & Visible & Green band; true-color composition (daytime) \\
\textbf{B03} & 0.64 & Visible & Cloud/scene detail; cloud amount (daytime) \\
B04 & 0.86 & NIR & Vegetation/reflectance; cloud phase aid \\
B05 & 1.6  & NIR & Cloud phase (ice vs. water); snow--cloud separation; hotspot aid \\
B06 & 2.3  & NIR & Cloud microphysics (particle size); hotspot aid \\
\textbf{B07} & 3.9  & IR (SWIR) & Night fog/low cloud; fires/hotspots; cloud-top temperature support \\
B08 & 6.2  & IR (WV) & Upper-tropospheric water vapor; jet/upper-level dynamics \\
B09 & 6.9  & IR (WV) & Mid-level water vapor; moisture structure \\
\textbf{B10} & 7.3  & IR (WV) & Lower-level water vapor; dry intrusion/convection environment \\
B11 & 8.6  & IR & Cloud phase/microphysics; ash/SO$_2$ discrimination aid \\
B12 & 9.6  & IR (O$_3$) & Ozone absorption; stratospheric influence; deep convection top features \\
B13 & 10.4 & IR window & Primary cloud-top brightness temperature; cloud-top height proxy \\
\textbf{B14} & 11.2 & IR window & Split-window combinations for fog/dust/ash; microphysics \\
B15 & 12.4 & IR window & Split-window for fog/low cloud, dust; SST/LST retrieval support \\
B16 & 13.3 & IR (CO$_2$) & CO$_2$ slicing for cloud-top height; thin cirrus detection \\
\hline
\end{tabular}
\end{table*}

\begin{table*}[h]
\caption{Variables from ERA5 and Baguan used for training and inference in Baguan-solar.}
\label{tab:dataset_baguan}
\begin{center}
% \begin{tiny}
% \usepackage{multirow}

\begin{tabular}{llccccc}
\toprule
Type & Variable name & Abbrev. & Stage1 input & Stage2 input & Levels \\
% Static & geopotential at surface & - & $\checkmark$ & $\times$ & 162051 & -\\

\midrule
% Single & 2 metre temperature & T2m & $\checkmark$ & $\checkmark$ & - \\
% Single & 10 metre U wind component & U10 & $\checkmark$ & $\checkmark$ & - \\
% Single & 10 metre V wind component & V10 & $\checkmark$ & $\checkmark$ & - \\
Single & low cloud cover & LCC & $\checkmark$ &  & - \\
Single & total cloud cover & TCC & $\checkmark$ &  & - \\
Single & total column water & TCW & $\checkmark$ & $\checkmark$ & - \\
Single & total column water vapour & TCWV & $\checkmark$ & $\checkmark$ & - \\
Single & total sky direct solar radiation at surface & FDIR & $\checkmark$ & $\checkmark$ & - \\
Single & surface solar radiation downwards & SSRD & $\checkmark$ & $\checkmark$ & - \\
\midrule

Atmospheric & U wind component & U & $\checkmark$ & & 50, 250, 500, 600, 700, 850, 925 \\
Atmospheric & V wind component & V & $\checkmark$ &  & 50, 250, 500, 600, 700, 850, 925 \\
Atmospheric & Temperature & T & $\checkmark$ &  & 50, 250, 500, 600, 700, 850, 925 \\
Atmospheric & Specific humidity & Q & $\checkmark$ & $\checkmark$ & 50, 250, 500, 600, 700, 850, 925 \\
Atmospheric & Geopotential & Z & $\checkmark$ & & 50, 250, 500, 600, 700, 850, 925 \\

% \midrule
% Atmospheric & U wind component & U & $\checkmark$ & $\checkmark$& 131 & 50, 100, 150, 200, 250, 300, 400, 500, 600, 700, 850, 925, 1000 \\
% Atmospheric & V wind component & V & $\checkmark$ & $\checkmark$& 132 & 50, 100, 150, 200, 250, 300, 400, 500, 600, 700, 850, 925, 1000 \\
% Atmospheric & Temperature & T & $\checkmark$ & $\checkmark$& 130 & 50, 100, 150, 200, 250, 300, 400, 500, 600, 700, 850, 925, 1000 \\
% Atmospheric & Specific humidity & Q & $\checkmark$ & $\checkmark$& 133 & 50, 100, 150, 200, 250, 300, 400, 500, 600, 700, 850, 925, 1000 \\
% Atmospheric & Relative humidity & R & 157 & 50, 250, 500, 600, 700, 850, 925 \\
\bottomrule
\end{tabular}

\end{center}
\end{table*}

\subsection{Limitations of weather foundation models}

Table~\ref{app:tab:compare_wfms} highlights a key gap in current weather foundation models (WFMs): despite their strong performance in general meteorological forecasting, large-scale AI models that explicitly predict cloud cover or solar irradiance remain largely unavailable. Baguan-solar is designed to bridge this gap by extending WFMs toward cloud-aware, day-ahead solar irradiance forecasting.

\subsection{Himawari-8/9 Satellite Data}

Table~\ref{app:tab:himawari_bands} summarizes the 16 spectral bands of the Himawari-8/9 Advanced Himawari Imager (AHI), including their center wavelengths, band types (visible, near-infrared, and infrared), and typical meteorological applications. These channels provide complementary information on cloud amount and texture in the visible range, cloud phase and microphysics in the near-infrared, and cloud-top temperature/height as well as water-vapor structure in the infrared. As described by JMA/MSC (2024)~\cite{JMA_MSC_UtilizationMetSatData}, Band~3 (0.64~$\mu$m) measures reflected visible solar radiation and supports true-color composites as well as daytime identification of low clouds and fog. Band~7 (3.9~$\mu$m) senses emitted terrestrial radiation and includes a substantial reflected solar component during daytime; at night, it supports hotspot detection and fog/low-cloud identification via the Band~7--Band~13 brightness temperature difference. Band~10 (7.3~$\mu$m) is a water vapor channel primarily sensitive to mid-tropospheric moisture and can also respond to volcanic SO$_2$. Band~14 (11.2~$\mu$m) is a longwave infrared window channel used for cloud imaging and cloud-top characterization, and it can support surface temperature applications under clear-sky conditions.

\subsection{Hyperparameter details}
\label{app:sec:hyperparams}
Baguan-solar uses a two-stage Swin Transformer design. The following list summarizes the hyperparameters used for Baguan-solar.

\begin{lstlisting}[caption={Hyperparameters of Baguan-solar.}, label=lst:baguan_solar_hparams]
image_size: [512, 512]
patch_size: [8, 8]
window_size: 16
embed_dim: 256
num_heads: [2]
patch_norm: True
drop_path_rate: 0.1
mlp_ratio: 4
qkv_bias: True
EnvEncoderSwinNet: 
  in_chans: 1170,
  out_chans: 256,
  depths: [8]
SateEncoderSwinNet: 
  in_chans: 24,
  out_chans: 256,
  depths: [8]
MultiDecoderSwinNet_Stage1: 
  in_chans: 512,
  out_chans: 120,
  depths: [2]
MultiDecoderSwinNet_Stage2: 
  in_chans: 17,
  out_chans: 1,
  depths: [8]

\end{lstlisting}

\subsection{Fast Vectorized Clear-Sky GHI Computation}
\label{app:sec:fast_ghi}
Clear-sky global horizontal irradiance (GHI) provides an upper bound of surface irradiance under cloud-free conditions and is widely used as a physics-based prior for solar forecasting. In this work, we adopt the Ineichen--Perez clear-sky model~\cite{ineichen2002new} (as implemented in \texttt{pvlib}~\cite{holmgren2018pvlib}) but re-implement the critical steps using a lightweight, vectorized NumPy routine to enable high-throughput gridded computation.

Our forecasting pipeline requires clear-sky GHI on a dense spatial grid of size $512 \times 512$ for each time stamp. The \texttt{pvlib} library is primarily designed for site-based (per-location) clear-sky computations, necessitating an outer loop over grid cells for grid-wide evaluation. A direct call to \texttt{pvlib.clearsky.ineichen} introduces substantial per-point overhead, taking roughly 4 minutes per time step on a 512×512 grid with the standard \texttt{pvlib} pipeline.

To eliminate this bottleneck, we implement a streamlined clear-sky routine that (i) computes the solar zenith angle using a compact approximation inspired by the NREL Solar Position Algorithm (SPA), and (ii) rewrites the \texttt{pvlib.clearsky.ineichen} computation with fully vectorized NumPy broadcasting, allowing latitude, longitude, and elevation to be provided as 2D arrays and yielding clear-sky GHI over the entire raster in a single pass. On a $512\times512$ grid, this reduces the wall-clock time from ~4 minutes to ~1 second per time step (a $\sim240\times$ speedup), while preserving the original physical assumptions and keeping the numerical discrepancy within <1\%. The Ineichen–Perez clear-sky GHI is computed as follows:

Given the solar zenith angle \(z\) (degrees; determined by longitude, latitude, and time), site elevation \(h\) (meters), Linke turbidity \(T_L\) (dimensionless), day of year \(\mathrm{DOY}\), and air mass \(\mathrm{AM}\), the Ineichen–Perez clear-sky GHI \(I_{\mathrm{clear}}\) is computed as:
\begin{equation}
\label{eq:ineichen}
I_{\mathrm{clear}}
=
c_{g1}\, I_0 \cos(z)\,
\exp\!\Big(-c_{g2}\,\mathrm{AM}\,[f_{h1}+f_{h2}(T_L-1)]\Big)\,
\exp\!\big(0.01\,\mathrm{AM}^{1.8}\big),
\end{equation}
where \(I_0\) is the extraterrestrial irradiance (top-of-atmosphere normal irradiance), approximated by:
\begin{equation}
\label{eq:I0}
I_0 = 1367.7 \times \Big(1 + 0.033 \times \cos\big(\frac{2\pi}{365} \times \mathrm{DOY}\big)\Big).
\end{equation}
The elevation-dependent coefficients are:
\begin{equation}
\label{eq:cg1}
c_{g1} = 5.09\times10^{-5}\,h + 0.868,
\end{equation}

\begin{equation}
\label{eq:cg2}
c_{g2} = 3.92\times10^{-5}\,h + 0.0387,
\end{equation}

\begin{equation}
\label{eq:fh1}
f_{h1} = \exp(-h/8000),
\end{equation}

\begin{equation}
\label{eq:fh2}
f_{h2} = \exp(-h/1250).
\end{equation}

Air mass \(\mathrm{AM}\) is computed from \(z\) using a Kasten–Young–type approximation:
\begin{equation}
\label{eq:airmass}
\mathrm{AM}
=
\frac{1}{\cos(z)}.
\end{equation}

In our experiments, the proposed vectorized implementation makes clear-sky priors computationally practical at scale, enabling their use during both training and inference for long-horizon forecasting.

\end{document}